\documentclass[10pt,twocolumn,letterpaper]{article}

\usepackage{cvpr}
\usepackage{times}
\usepackage{epsfig}
\usepackage{graphicx}
\usepackage{amsmath}
\usepackage{amssymb}
\usepackage{setspace}
\usepackage{subcaption}

\usepackage[pagebackref=true,breaklinks=true,letterpaper=true,colorlinks,bookmarks=false]{hyperref}

\cvprfinalcopy 

\def\cvprPaperID{****} 

\ifcvprfinal\fi
\begin{document}


\title{Deep Variational Inference Without Pixel-Wise Reconstruction}

\author{Siddharth Agrawal and Ambedkar Dukkipati\\
Department of Computer Science and Automation\\
Indian Institute of Science\\
{\tt\small \{siddharth.agrawal, ad\}@csa.iisc.ernet.in}
}

\maketitle

\begin{abstract}
   Variational autoencoders (VAEs), that are built upon deep neural networks have
   emerged as popular generative models in computer vision. Most of the work towards
   improving variational autoencoders has focused mainly on making the approximations
   to the posterior flexible and accurate, leading to tremendous progress. However,
   there have been limited efforts to replace pixel-wise reconstruction, which
   have known shortcomings. In this work, we use real-valued non-volume preserving
   transformations (real NVP) to exactly compute the conditional likelihood of the
   data given the latent distribution. We show that a simple VAE with this form of
   reconstruction is competitive with complicated VAE structures, on image modeling
   tasks. As part of our model, we develop powerful conditional coupling layers that
   enable real NVP to learn with fewer intermediate layers.
\end{abstract}

\section{Introduction}

In recent years, variational autoencoders (VAEs)~\cite{VAE_1, VAE_2} have become
extremely popular for many machine learning problems. They have been used for a
variety of applications such as image modeling~\cite{ConceptCompress, DRAW},
interpretable representation learning~\cite{DC_IGN}, conditional image
generation~\cite{SemiSuperVAE, AlignDRAW} and 3D structure learning from images
\cite{ShapeNet}. VAEs provide a mathematically sound framework for unsupervised
learning by optimizing the variational lower bound on the data likelihood. This
lower bound involves two terms, (i) KL divergence of the approximate posterior with
a fixed prior and (ii) conditional likelihood of the data given the latent
distribution (also known as `reconstruction'). Much of the work in improving
VAEs has focused on modifying the approximate posterior for better expressivity
and approximations~\cite{ImpWeightAuto, IAF, NormFlow, MCMCVar}. On the other hand,
little work has been done to improve upon the form of the reconstruction. Most
VAE models assume a standard normal distribution for pixels in the
reconstructed image space that leads to a mean-squared reconstruction cost.
This has been previously shown to cause blurriness in the reconstructed images.
Previous work has attempted to circumvent the problem by augmenting the model with
generative adversarial networks~\cite{BeyondPix, BeyondMSE}. However, these models
do not allow one to compute the conditional likelihood term exactly which limits
our ability to objectively compare them with other VAE models. Other models have
used alternatives like discrete softmax distribution~\cite{PixelRNN} and discretized
logistic distribution~\cite{IAF} for pixels, but these have not been well studied
on their own.

Real-valued non-volume preserving transformations (real NVP)~\cite{NVP} offer
exact likelihood computation through non-linear invertible transformations whose
Jacobian determinants are easy to compute. This model also provides exact
inversion from the latent space to the data space enabling efficient sampling,
which is not available in other exact likelihood methods such as pixel recurrent
neural networks (Pixel RNN)~\cite{PixelRNN} and pixel convolutional neural
networks (Pixel CNN)~\cite{PixelCNN}. We use real NVP transformations to exactly
compute the conditional likelihood term in a VAE and thus alleviate the problem
of mean-squared reconstruction. We show that just using this modification we can
compete with other complicated VAE models such as convolutional DRAW~\cite{ConceptCompress}
(which uses multiple stochastic layers and recursion for sample generation) as
well as real NVP~\cite{NVP} (using a smaller architecture). A summary of our
contributions is as follows:

(1) We propose a model that uses real NVP transformations to model the conditional
likelihood of the data given the latent distribution in a VAE.

(2) We propose a conditional coupling layer to make conditioning on the latent
distribution stronger, adding multiplicative interactions to enable expressivity
in the model with fewer layers.

(3) We compare the model against a complicated VAE model (convolutional DRAW) and
also against other state-of-the-art generative models.

In the following section we review some of the preliminaries, and then go on to
describe our model formally.

\section{Background}

\subsection{Variational autoencoder}
Variational autoencoders differ from regular autoencoders in that they have one
or more stochastic layers for latent variables. These latent variables form
the approximate posterior $q(z|x)$, which is forced to be close to
a chosen prior such as a standard normal distrbution. This is achieved
by minimizing the KL divergence between the approximate posterior and the prior,
one of two terms in the variational lower bound for the log-likelihood of the
data~\cite{VAE_1}:
\begin{align}
   \text{log}(p(x)) \geq \mathbb{E}_{q(z|x)}[\text{log}(p(x|z))] - \text{KL}(q(z|x)||p(z))
\end{align}
Here, $q(z|x)$ is the approximate posterior modeled by the encoder, $p(x|z)$ is
the conditional likelihood modeled by the decoder and $p(z)$ is the fixed prior
distribution. In an unsupervised learning setup, one maximizes the
variational lower bound as a surrogate for the log-likelihood. The expectation
term is estimated using Monte Carlo sampling over the batch,
$\mathbb{E}_{q(z|x)}[\text{log}(p(x|z))] \approx \frac{1}{n}\sum_{i=1}^n \text{log}(p(x_i|z_i))$,
where $x_i, i = 1, 2, \dots, n$ is a training example in the batch. In this work, we improve upon
the technique to calculate $p(x_i|z_i)$. Instead of assuming that the reconstructed
image space follows a standard normal distribution, we assume that an intermediate
layer follows a parametrized normal distribution. We provide more details in the
next section.

\subsection{Real NVP}
Real NVP~\cite{NVP} is an exact likelihood model, that transforms the data into a prior
probability distribution. Let us say that the data space $X$ is transformed into
the space $Y$ through the function $f$. The change of variable formula for this
transformation is given by the following equation:
    \begin{align}
        p_X(x) &= p_Y(f(x)) \left|\text{det}\left(\frac{\partial f(x)}{\partial x^T}\right)\right|
    \end{align}
    Here, $x$ is a point in the data space. From the above equation, the likelihood
of the data can be estimated if we can compute the two terms on the right. The
likelihood of $f(x)$ in the space $Y$ can be computed analytically if we assume a
prior such as a standard normal distribution in that space. To compute the second term,
we need to be able to calculate the determinant of the Jacobian of the transformation
$f(x)$. As we will see, this is enabled by the coupling layer transform. Also, if
$f(x)$ is invertible, we can easily go from the latent space to the
data space, $x = f^{-1}(y)$.

Let $x$ be the input to the coupling layer and $y$ be the expected output, each
of the vectors are assumed to be $D$ dimensional. The coupling layer transform is
given by \cite{NVP}:
    \begin{align}
        y_{1:d} &= x_{1:d}\\
        y_{d+1:D} &= x_{d+1:D} \odot \text{exp}(l(x_{1:d})) + m(x_{1:d})
    \end{align}
    Here, $l$ and $m$ are arbitrary functions, which can be realized using neural
networks. The transform divides the input vector into two parts, where the first
part is kept unchanged and the second part is transformed using a function of the
unchanged part.

The coupling layer transform can be inverted using the following equations:
    \begin{align}
        x_{1:d} &= y_{1:d}\\
        x_{d+1:D} &= (y_{d+1:D} - m(y_{1:d})) \odot \text{exp}(-l(y_{1:d}))
    \end{align}
    The computational complexity of computing the inverse is the same as going in
the forward direction. If $f(x)$ is taken as a series of coupling layers, then one
can compute $f^{-1}$ as each of the coupling layers is invertible.

The Jacobian of the coupling layer transform is as follows:
    \begin{align}
        \frac{\partial y}{\partial x^T} &=
        \begin{bmatrix}
            \mathbb{I}_d & 0 \\
            \frac{\partial y_{d+1:D}}{\partial x_{1:d}^T} & \text{diag}(\text{exp}(l(x_{1:d})))
        \end{bmatrix}
    \end{align}
    Here, $\mathbb{I}_d$ is a $d$-dimensional identity matrix. The transformation
gives a triangular Jacobian whose determinant is very easy to compute. From the
above equation,
    \begin{align}
        \text{det}\left(\frac{\partial y}{\partial x^T}\right) = \text{exp}\left(\sum l(x_{1:d})\right)
    \end{align}
    The summation is over the components of $l(x_{1:d})$.
    One can see the effect of a series of coupling layer transforms on the determinant
of the Jacobian. If the data is transformed from $x$ to $y_1$, and from $y_1$ to $y_2$,
one can compute the determinant as follows:
    \begin{align}
        \text{det}\left(\frac{\partial y_2}{\partial x^T}\right) &= \text{det}\left(\left(\frac{\partial y_2}{\partial y_1^T}\right) \left(\frac{\partial y_1}{\partial x^T}\right)\right)\nonumber\\
        &= \text{det}\left(\frac{\partial y_2}{\partial y_1^T}\right) \text{det}\left(\frac{\partial y_1}{\partial x^T}\right)
    \end{align}
    This shows that one can compute the determinant of the Jacobian of $f(x)$ taken
as a series of coupling layers by just multplying determinants of the Jacobians of
the individual coupling layers. This enables us to compose $f(x)$ using an arbitrary number
of coupling layers. Because of the above mentioned properties of the coupling layer
transform, we can now compute the exact likelihood of the data. Thus, the network
can be optimized using the maximum likelihood framework.

\begin{figure*}
   \begin{center}
      \includegraphics[width=0.6\linewidth]{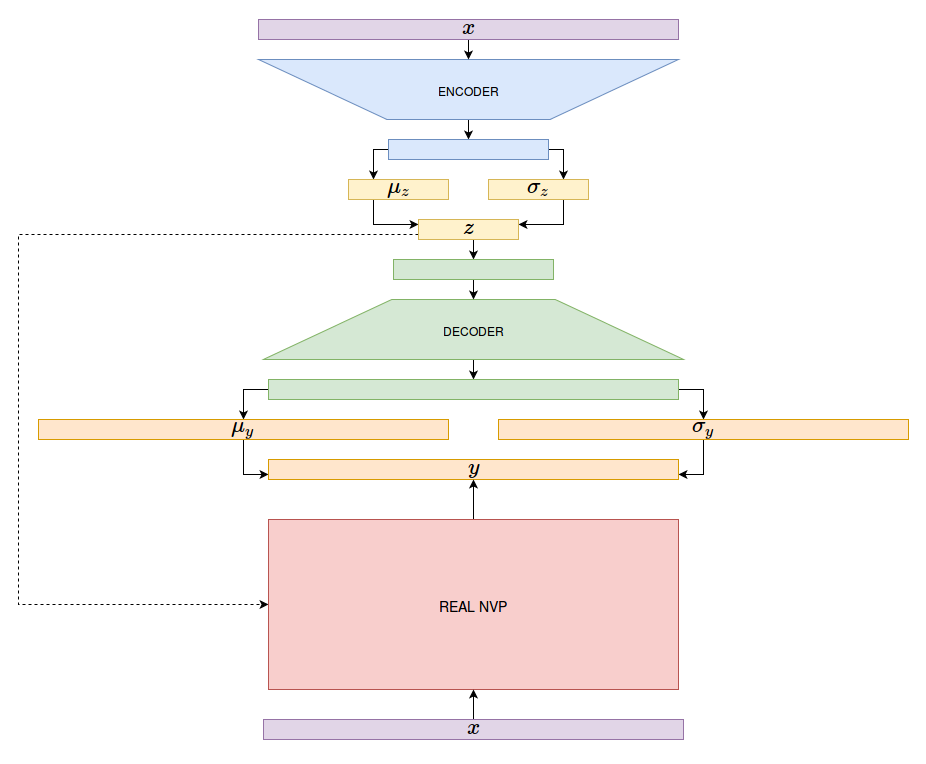}
   \end{center}
   \caption{Block diagram of VAPNEV, color coded to indicate independent components.
            The dotted line suggests that the latent distribution may or may not
            be used for conditioning in the coupling layer transforms. The figure
            is illustrative of relative network sizes.}
   \label{fig:arch_block}
\end{figure*}

\section{Our model: VAPNEV}

In this section, we formally describe our model with the architectural novelties
that we use. We call our model VAPNEV, which is an anagram for VAE-NVP.

\subsection{Conditional likelihood calculation}
In a regular VAE, the reconstructed image space is assumed to follow a normal
distribution. Instead, here we assume an intermediate space $Y$ to follow a normal distribution.
In order to calculate $p(x|z)$, we can transform the data space $X$ into the intermediate
space $Y$ that depends on VAE latent space $Z$. The change of variable formula for this
transformation is:
\begin{align}
   p(x|z) &= p(f_z(x)|z) \left|\text{det}\left(\frac{\partial f_z(x)}{\partial x^T}\right)\right|
\end{align}
Here, $f_z(x)$ is a function that projects from the space $X$ to the space $Y$.
In general, the transformation can depend on $z$ indicated by the subscript. If we
assume $Y \sim \mathcal{N}(\mu_y, \Sigma_y)$, where $\mu_y = \phi_1(z)$ and
$\Sigma_y = \text{diag}(\phi_2(z))$, log($p(f_z(x)|z)$) can be calculated using:
\begin{align}
    \text{log}(p(f_z(x)|z)) =& -\frac{1}{2}\text{log}(|\Sigma_y|)\nonumber\\
                             & -\frac{1}{2} (f_z(x) - \mu_y)^T \Sigma_y^{-1} (f_z(x) - \mu_y)\nonumber\\
                             & -\frac{D}{2}\text{log}(2\pi)
\end{align}
where $D$ is the dimensionality of the space. The determinant of the Jacobian
$\left|\text{det}\left(\frac{\partial f_z(x)}{\partial x^T}\right)\right|$ can
be computed if $f_z(x)$ is taken to be a series of coupling layers, as seen in the
previous section. Thus, using this formulation with real NVP, one can completely avoid
pixel-wise computation and still exactly calculate the conditional likelihood. It
should be noted that the formulation holds even if $f_z(x)$ does not depend on $z$. As
can be seen in Figure~\ref{fig:arch_block}, VAPNEV models the prior space of the
NVP using the decoder output.

\begin{figure}[t]
   \begin{center}
      \includegraphics[width=0.8\linewidth]{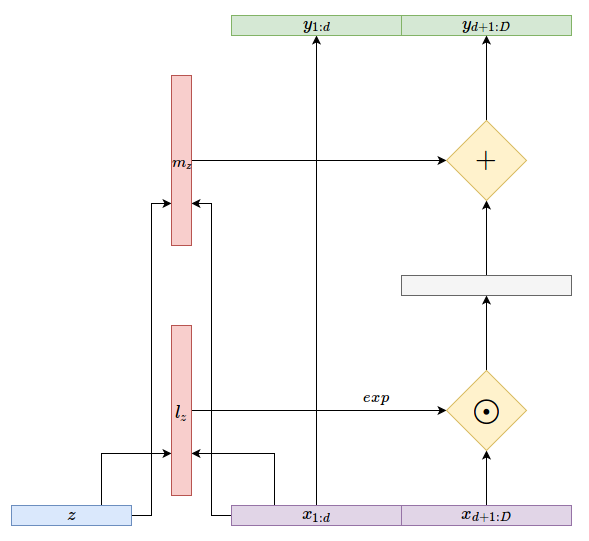}
   \end{center}
   \caption{Conditional coupling layer}
   \label{fig:cond_coup}
\end{figure}

\subsection{Conditional coupling layer}
In order to make $f_z(x)$ conditional on the latent distribution of the VAE we propose the
conditional coupling layer. This layer satisfies the following two conditions
which are necessary for it to be useful in the above scenario, (i) the determinant
of the Jacobian should be easy to compute and (ii) it should be invertible given
$z$. The first condition ensures efficient computation of the cost of the model,
whereas the second condition is essential for efficient sampling. The conditional
coupling layer transform is very similar to the original transform and is given by:
\begin{align}
    y_{1:d} &= x_{1:d}\\
    y_{d+1:D} &= x_{d+1:D} \odot \text{exp}(l_z(x_{1:d})) + m_z(x_{1:d})
\end{align}
Here, $x$ and $z$ are input to the layer, and $y$ is the output of the layer, $x$
and $y$ being $D$-dimensional. $l_z$ and $m_z$ can be arbitrary functions dependent
on $z$. As for the regular coupling layers, this transformation gives us a diagonal
Jacobian, whose determinant is easy to compute. The inverse can also be computed
using very similar equations, provided $z$ is a known value:
\begin{align}
    x_{1:d} &= y_{1:d}\\
    x_{d+1:D} &= (y_{d+1:D}-m_z(y_{1:d})) \odot \text{exp}(-l_z(y_{1:d}))
\end{align}
A graphical representation of the conditional coupling layer is shown in
Figure~\ref{fig:cond_coup}. We now discuss the exact form of $l_z(x)$, which is also used for $m_z(x)$. From
the transformation equations it is apparent that $l_z(x)$ should project to
$\mathbb{R}^{D-d}$. To achieve this, we project each of $x$ and $z$ to $\mathbb{R}^{D-d}$
using functions $l_1$ and $l_2$ respectively, and then operate in an elementwise
fashion on the computed function values. Note that here $x \in \mathbb{R}^{d}$ and
if we take $d = D/2$, then $l_1$ can be a residual network~\cite{ResNet} that maintains the
dimensionality of $x$. Since $z$ is generally a low dimensional vector, we take
$l_2$ to be a deconvolution network. Inspired by \cite{MIRNN}, we use multiplicative
interactions between $l_1(x)$ and $l_2(z)$ to increase expressivity of the function.
We summarize $l_z(x)$ using the following equation:
\begin{align}
    l_z(x) &= \alpha \odot l_1(x) \odot l_2(z) + \beta_1 \odot l_1(x) + \beta_2 \odot l_2(z) + b
\end{align}
Since both $l_1(x)$ and $l_2(z)$ are outputs of convolutional networks, they are
tensors in $\mathbb{R}^{H\text{x}W\text{x}C}$ ($H\text{x}W\text{x}C = D-d$). The
multipliers $\alpha, \beta_1, \beta_2$ and the bias $b$ are all in $\mathbb{R}^C$,
and are trainable parameters. The elementwise multiplication with these multipliers
uses broadcasting. The conditional coupling layer allows for shortcut connections
to the VAE latent distribution which allows for stronger conditioning and faster
training.

\subsection{Training}
The model is trained to maximixe the variational lower bound on the data log-likelihood.
We summarize the feedforward computation of VAPNEV as shown in Figure~\ref{fig:arch_block}:
\begin{align}
    \mu_z &= \psi_1(x)\\
    \sigma_z &= \psi_2(x)\\
    z \sim q(z|x) &= \mathcal{N}(\mu_z, \sigma_z)\\
    \mu_y &= \phi_1(z)\\
    \sigma_y &= \phi_2(z)\\
    y &= f_z(x)
\end{align}
$\psi_1$ and $\psi_2$ include the encoder as well as the respective projections
to the mean and variance of the approximate posterior. $z$ is sampled from the
approximate posterior using the reparametrization trick~\cite{VAE_1}. $\phi_1$ and $\phi_2$
include the decoder as well as the respective projections to the mean and variance
of the NVP latent space. $f_z(x)$ is the NVP network consisting of conditional
coupling layers. Using $\mu_z$ and $\sigma_z$, the KL divergence term can
be computed since we assume $p(z) = \mathcal{N}(0, I)$. $y$, $\mu_y$ and $\sigma_y$
are used to calculate $\text{log}(p(f_z(x)|z))$ and the adjustment
$\text{log}\left(\left|\text{det}\left(\frac{\partial f_z(x)}{\partial x^T}\right)\right|\right)$
is provided by $f_z(x)$, which gives us $\text{log}(p(x|z))$.

We face the same problem as in \cite{SentVAE}, where the KL divergence term quickly
goes to zero, leading to undesirable local optima. We use the annealing procedure
suggested in \cite{SentVAE} to alleviate the issue. We optimize our model using
the ADAM optimizer~\cite{ADAM} with default hyperparameters.

\subsection{Generation}
The generation process for VAPNEV is straightforward, which we summarize using
the following equations:
\begin{align}
    z \sim p(z) &= \mathcal{N}(0, I)\\
    \mu_y &= \phi_1(z)\\
    \sigma_y &= \phi_2(z)\\
    y &\sim \mathcal{N}(\mu_y, \sigma_y)\\
    x &= f_z^{-1}(y)
\end{align}
Unlike a regular VAE, a single $z$ might lead to different samples in VAPNEV,
because of stochasticity in the $Y$ space. In case this is not desirable, we can
pass $\mu_y$ into $f_z^{-1}$. We are able to calculate $f_z^{-1}(y)$
because of the invertibility property of the conditional coupling layer given $z$.

\subsection{Reconstruction}
Reconstruction of a given batch is very similar to the generative process, the
only difference being that $z$ is sampled from the approximate posterior instead
of the prior:
\begin{align}
    \mu_z &= \psi_1(x)\\
    \sigma_z &= \psi_2(x)\\
    z \sim q(z|x) &= \mathcal{N}(\mu_z, \sigma_z)\\
    \mu_y &= \phi_1(z)\\
    \sigma_y &= \phi_2(z)\\
    y &\sim \mathcal{N}(\mu_y, \sigma_y)\\
    x &= f_z^{-1}(y)
\end{align}
The inverse of the NVP network can be seen as an extension of the decoder, as it
is the final decoding step in the generation and reconstruction steps.

\begin{figure*}
   \hspace*{\fill}
   \begin{subfigure}{0.3\textwidth}
      \centering
      \includegraphics[width=\linewidth]{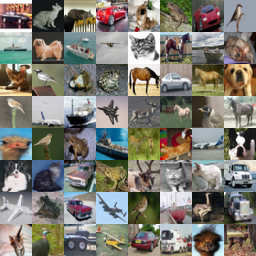}
   \end{subfigure}
   \hfill
   \begin{subfigure}{0.3\textwidth}
      \centering
      \includegraphics[width=\linewidth]{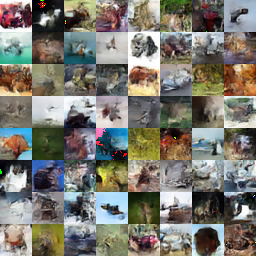}
   \end{subfigure}
   \hfill
   \begin{subfigure}{0.3\textwidth}
      \centering
      \includegraphics[width=\linewidth]{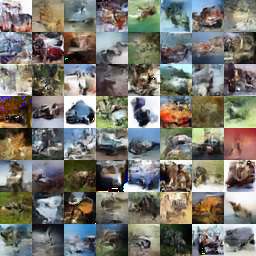}
   \end{subfigure}
   \hspace*{\fill}
   \caption{The left panel shows original images in the CIFAR-10 dataset. The
            middle panel shows the reconstructions of the images in the left panel.
            The right panel contains images sampled randomly from the model.}
   \label{fig:cifar}
\end{figure*}

\begin{figure*}
   \hspace*{\fill}
   \begin{subfigure}{0.3\textwidth}
      \centering
      \includegraphics[width=\linewidth]{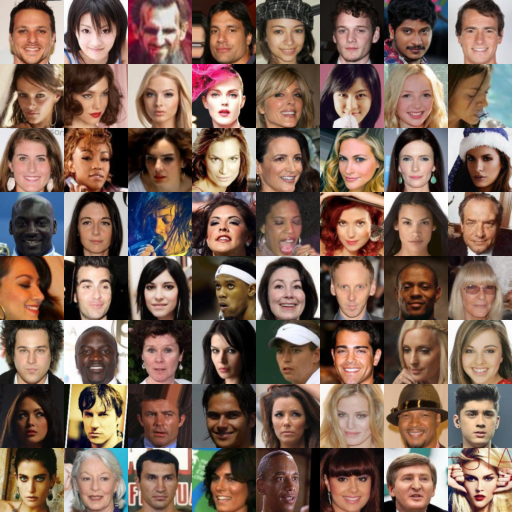}
   \end{subfigure}
   \hfill
   \begin{subfigure}{0.3\textwidth}
      \centering
      \includegraphics[width=\linewidth]{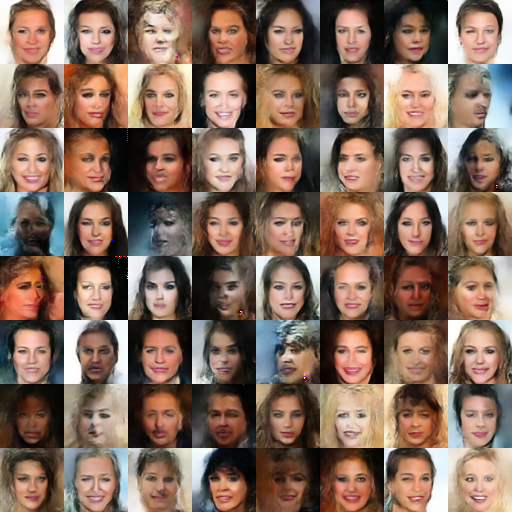}
   \end{subfigure}
   \hfill
   \begin{subfigure}{0.3\textwidth}
      \centering
      \includegraphics[width=\linewidth]{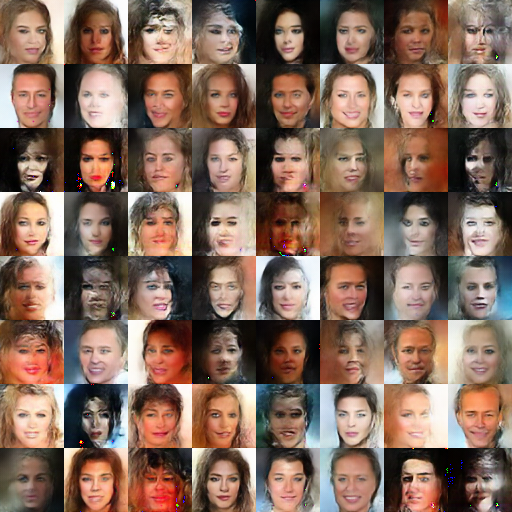}
   \end{subfigure}
   \hspace*{\fill}
   \caption{The left panel shows original images in the CelebA dataset. The
            middle panel shows the reconstructions of the images in the left panel.
            The right panel contains images sampled randomly from the model.}
   \label{fig:celeb}
\end{figure*}

\section{Experimental results}
We test our model on the task of generative image modeling. The model is compared
using natural images (CIFAR-10~\cite{CIFAR}), as well as fixed domain images
(CelebA~\cite{CelebA}). First, we specify some common details used in all the
experiments.

\subsection{Modeling transformed data}
The discrete image data in $[0, 255]^D$ is first corrupted with uniform noise in
$[0, 1]$ to make it continuous, and is then scaled to $[0, 1]^D$. Since real NVP
gives a transformation from $\mathcal{R}^D$ to $\mathcal{R}^D$, we model the density
of $\text{log}(\frac{x'}{1-x'})$~\cite{NVP}, which takes $x'$ from $[0, 1]^D$ to
$\mathcal{R}^D$. Here, $x' = \alpha + (1-\alpha) \odot x, \alpha \in [0, 1]$,
which is done to avoid numerical errors within the log. In our experiments, we
take $\alpha = 0.05$. To compute the actual variational lower bound, we have to
account for this transformation. The correction factor comes out to be
$\sum \text{log}\left(\frac{1-\alpha}{x' \odot (1-x')}\right)$, where the sum
is over the components of $x'$. We also use horizontal flips of the dataset images
as data augmentation.

\subsection{Architectural details}
The encoder is taken as an 8-layer convolutional neural network. Every alternate
layer doubles the number of filters and halves the spatial resolution
in both directions. We start with 32 filters in the first layer. Analogous to the
encoder, the decoder is an 8-layer deconvolution network, that doubles the spatial
resolution and halves the number of filters every alternate layer. The first layer
of this deconvolution network starts with the same dimensions as the output of the last encoder
layer. The mean and variance of the VAE latent space are computed using separate
fully-connected linear projections of the encoder output; the dimensionality of
the latent space is taken to be 256. In case of the NVP latent space, the mean and
variance are separate convolutional linear projections.

For the conditional coupling layer transform, we use checkerboard masking, channel-wise
masking and the squeeze operation, all mentioned in \cite{NVP}. To compute $l_z(x)$,
we take $l_1(x)$ as a network of 2 residual blocks and $l_2(z)$ as a small
deconvolution network. This deconvolution network starts with a tensor of
$2 \times 2 \times c$, and doubles the spatial resolution at each layer. We take
$c$ to be the number of filters in the residual blocks of the coupling layer. The
same configuration is used for $m_1(x)$ and $m_2(z)$ in $m_z(x)$. We use a
multi-scale architecture as mentioned in \cite{NVP}, with 2 scales. Each scale
has 3 conditional coupling layers with checkerboard masking and 3 with channel-wise
masking. We start with 64 filters for residual blocks in the first scale, and double
them for the next scale. We use the same architecture for both the datasets.

\begin{table}
  \begin{center}
    \begin{tabular}{|l|c|}
      \hline
      Method & Bits/dim \\
      \hline\hline
      PixelRNN~\cite{PixelRNN} & 3.00 \\
      IAF~\cite{IAF} & $<$ 3.28 \\
      Real NVP~\cite{NVP} & 3.49\\
      Conv DRAW~\cite{ConceptCompress} & $<$ 3.59\\
      VAPNEV & $<$ 3.55\\
      \hline
    \end{tabular}
  \end{center}
  \caption{Bits/dim on CIFAR-10. Lower is better.}
  \label{tab:cifar}
\end{table}

\subsection{CIFAR-10}
From Table~\ref{tab:cifar}, we can see that VAPNEV is competitive with convolutional
DRAW which is a complicated VAE structure with multiple stochastic layers and
recurrent feedback. This establishes that replacing pixel-wise reconstruction with
exact likelihood methods like real NVP is beneficial to the performance of VAEs.
The model is also competitive with real NVP, which uses a much bigger architecture
(8 residual blocks in coupling layers as opposed to 2 here). This shows the power
of the conditional coupling layer transform, which is able to effectively utilize the
semantic representation learned by the VAE latent distribution. As can be seen from
the reconstructions in Figure~\ref{fig:cifar}, VAPNEV learns to model high level
semantics in the latent distribution such as background color, pose and location
of the object. The samples also show that the model is able to learn better global
structure.

\begin{table}
  \begin{center}
    \begin{tabular}{|l|c|}
      \hline
      Method & Bits/dim \\
      \hline\hline
      Real NVP~\cite{NVP} & 3.02\\
      VAPNEV & $<$ 2.8\\
      \hline
    \end{tabular}
  \end{center}
  \caption{Bits/dim on CelebA. Lower is better.}
  \label{tab:celeb}
\end{table}

\subsection{CelebA}
As shown in Table~\ref{tab:celeb}, VAPNEV performs significantly better than NVP
on CelebA, while having a smaller architecture (The NVP model has 4 scales for
CelebA, whereas VAPNEV uses 2). This suggests that NVP can be improved by
using better global representations, learned here by the VAE. Looking at the
reconstructions in Figure~\ref{fig:celeb}, we can see that the model learns high
level semantic features such as hair color, face pose and expressions.

\section{Discussion and future work}
In this paper, we suggest a way to replace pixel-wise reconstruction with a maximum
likelihood based alternative. We show that this greatly benefits the VAE formulation,
as a simple VAE augmented with NVP transformations is able to compete with complicated
models with multiple stochastic layers and recurrent connections. We develop powerful
conditional coupling layer transforms which enable the model to learn with smaller
architectures. VAPNEV provides a lot of advantages such as (i) it provides a way
to replace pixel-wise reconstruction which has known shortcomings, (ii) it gives
a generative model which can be trained and sampled from efficiently and (iii) it
is a latent variable model which can be used for downstream supervised or semi-supervised
learning.

This work can be extended in several ways. Using deeper architectures, and combining
with expressive posterior computations like inverse autoregressive flow~\cite{IAF},
it may be possible to compete with or even beat state-of-the-art models. This technique
can be used to improve VAE models for other tasks such as semi-supervised learning
and conditional density modeling. The conditional coupling layer can be used for
constructing conditional real NVP models.

{\small
\bibliographystyle{ieee}
\bibliography{egbib}
}

\end{document}